\documentclass[iicol,sn-basic]{sn-jnl}% Style for submissions to Nature Portfolio journals
\usepackage{graphicx}%
\usepackage{multirow}%
\usepackage{amsmath,amssymb,amsfonts}%
\usepackage{amsthm}%
\usepackage{mathrsfs}%
\usepackage[title]{appendix}%
\usepackage{xcolor}%
\usepackage{textcomp}%
\usepackage{manyfoot}%
\usepackage{booktabs}%
\usepackage{algorithm}%
\usepackage{algorithmicx}%
\usepackage{algpseudocode}%
\usepackage{listings}%
\usepackage{tabularx}
\usepackage{natbib}
\theoremstyle{thmstyleone}%
\theoremstyle{thmstyletwo}%

\theoremstyle{thmstylethree}%

\begin{document}

\title[Article Title]{ETTrack: Enhanced Temporal Motion Predictor for Multi-Object Tracking}

%%=============================================================%%
%% GivenName	-> \fnm{Joergen W.}
%% Particle	-> \spfx{van der} -> surname prefix
%% FamilyName	-> \sur{Ploeg}
%% Suffix	-> \sfx{IV}
%% \author*[1,2]{\fnm{Joergen W.} \spfx{van der} \sur{Ploeg} 
%%  \sfx{IV}}\email{iauthor@gmail.com}
%%=============================================================%%

\author{\fnm{Xudong} \sur{Han}}\email{xh218@sussex.ac.uk}
%%%%%%%%%%%%%%%%%%%%%%%%%%%%%%%%%%%%%%%%%%%%%%%%%%%%%%%%%
%\author{\fnm{Nobuyuki} \sur{Oishi}}\email{n.oishi@sussex.ac.uk}
%\author{\fnm{Yueying} \sur{Tian}}\email{yt322@sussex.ac.uk}
%\equalcont{These authors contributed equally to this work.}
%\author{\fnm{Elif} \sur{Ucurum}}\email{e.ucurum@sussex.ac.uk}
%\author{\fnm{Rupert} \sur{Young}}\email{r.c.d.young@sussex.ac.uk}
%\author{\fnm{Chris} \sur{Chatwin}}\email{c.r.chatwin@sussex.ac.uk}
\author{\fnm{Nobuyuki} \sur{Oishi}}
\author{\fnm{Yueying} \sur{Tian}}
%\equalcont{These authors contributed equally to this work.}
\author{\fnm{Elif} \sur{Ucurum}}
\author{\fnm{Rupert} \sur{Young}}
\author{\fnm{Chris} \sur{Chatwin}}
%%%%%%%%%%%%%%%%%%%%%%%%%%%%%%%%%%%%%%%%%%%%%%%%%%%%%%%%%%%%%%%%%%%
\author*{\fnm{Philip} \sur{Birch}}\email{P.M.Birch@sussex.ac.uk}

%\equalcont{These authors contributed equally to this work.}
\affil{\orgdiv{School of Engineering and Informatic}, \orgname{University of Sussex}, \orgaddress{\city{Falmer}, \postcode{BN1 9QT}, \state{East Sussex}, \country{United Kingdom}}}

%%==================================%%
%% Sample for unstructured abstract %%
%%==================================%%

\abstract{Many Multi-Object Tracking (MOT) approaches exploit motion information to associate all the detected objects across frames. However, many methods that rely on filtering-based algorithms, such as the Kalman Filter, often work well in linear motion scenarios but struggle to accurately predict the locations of objects undergoing complex and non-linear movements. To tackle these scenarios, we propose a motion-based MOT approach with an enhanced temporal motion predictor, ETTrack. Specifically, the motion predictor integrates a transformer model and a Temporal Convolutional Network (TCN) to capture short-term and long-term motion patterns, and it predicts the future motion of individual objects based on the historical motion information. Additionally, we propose a novel Momentum Correction Loss function that provides additional information regarding the motion direction of objects during training. This allows the motion predictor rapidly adapt to motion variations and more accurately predict future motion. Our experimental results demonstrate that ETTrack achieves a competitive performance compared with state-of-the-art trackers on DanceTrack and SportsMOT, scoring 56.4$\%$ and 74.4$\%$ in HOTA metrics, respectively.}


%%\pacs[JEL Classification]{D8, H51}

%%\pacs[MSC Classification]{35A01, 65L10, 65L12, 65L20, 65L70}

\maketitle

\section{Introduction}\label{sec1}

Multi-Object Tracking (MOT) is an important technology in the field of computer vision and plays a significant role in applications such as mobile robotics\citep{yuan2022glamr}, autonomous driving\citep{sun2020scalability}, and sports analytics\citep{zhao2023survey}. With recent progress in object detection, tracking-by-detection methods\citep{bewley2016simple,wojke2017simple,du2023strongsort} have become the most popular paradigm. These approaches typically comprise two subtasks: detecting objects in each frame; and associating them across various frames. The core of the tracking-by-detection paradigm is data association, which relies heavily on the utilization of object appearance and motion information for accuracy. Despite the benefits of employing detection to gain semantic advantages, this reliance poses significant challenges in complex scenarios where objects exhibit similar appearances and object occlusions occur frequently. Therefore, ReID-based MOT methods\citep{wojke2017simple,du2023strongsort,fischer2023qdtrack}, which use a trained ReID model to extract the appearance features of objects for data association, suffer from degradation of the tracking performance under such conditions. By contrast, motion information becomes relatively reliable in scenarios plagued by  similarities in appearance, blurring, and occlusions. 

Notably, motion model based MOT approaches\citep{bewley2016simple,zhang2022bytetrack,cao2023observation} utilize a motion predictor to recognize spatial and temporal patterns, thereby predicting future object movements for object association. However, it is still challenging for motion predictors to predict object motion in complex scenarios, such as those involving non-linear movements, diverse poses, and severe occlusions. In this work, our aim is to develop a temporal motion predictor that increases the accuracy of object association and tracking performance.

As the main branch of motion model-based tracking, filtering-based methods widely use the Kalman Filter\citep{kalman1960contributions} as a motion predictor with the assumption of constant velocity in both the prediction and filtering processes. The Kalman Filter works well with linear motion, but inaccurately predicts object locations in complex non-linear motion situations. To overcome these limitations, deep-learning-based motion models have been applied to MOT. For example, \citep{milan2017online,kesa2021joint} adopt Recurrent Neural Networks (RNNs) to predict the object position based on the historical trajectories of objects by exploiting their sequence processing capabilities. \citep{chaabane2021deft} employs a Long Short-Term Memory-based (LSTM) motion model  to capture motion constraints by considering the motion information of objects as input. \citep{xiao2023motiontrack} proposes a Transformer-based motion model to capture long-range dependencies for modeling motions. However, these deep-learning-based approaches have two limitations. First, owing to their simple network structure, when utilized in MOT they are unable to effectively handle input sequences with high variability and have difficulty modeling complex and long-range temporal dependencies \citep{luo2018fast}. Second, current motion models only consider the historical trajectories of objects as inputs and lack the capacity to integrate reliable additional information, resulting in unreliable position prediction in complex and non-linear scenes. Nonetheless, integrating additional information is possible, such as the appearance features of objects\citep{kesa2021joint} and the interaction between objects\citep{han2023ort}. However, when significant occlusions and predefined actions are present in these scenes\citep{sun2022dancetrack}, the performance is compromised.

To mitigate the adverse effects of these limitations, this work proposes two main innovations. We introduce an enhanced temporal motion predictor that integrates a Temporal Transformer model\citep{Vaswani_Shazeer_Parmar_Uszkoreit_Jones_Gomez_Kaiser_Polosukhin_2017} and a Temporal Convolutional Network (TCN)\citep{bai2018empirical} for MOT. The successful application of the Transformer model in Natural Language Processing (NLP) domains\citep{Vaswani_Shazeer_Parmar_Uszkoreit_Jones_Gomez_Kaiser_Polosukhin_2017,devlin2018bert,lan2019albert} has demonstrated its capability to model long-range temporal dependencies using a powerful self-attention mechanism. We utilize a specialized Temporal Transformer architecture that utilizes only the encoder component of the traditional Transformer model. This design efficiently captures the historical motion patterns of individual objects to predict their motion. The TCN, which employ dilated causal convolution, models human motion to capture basic motion patterns and temporal interdependencies. Furthermore, the TCN can capture the motion patterns of objects on various time scales, particularly short-term minor changes, through the adaptable adjustment of the convolutional kernel size and expansion factor. Integrating the Temporal Transformer and the TCN enables our motion predictor to comprehend both local and global motion information. The TCN excels at capturing fine-grained motion details, while the Temporal Transformer builds upon this to comprehend broader, long-term motion trends.

In addition to the motion predictor, we study the influence of the motion direction of the objects on their predicted positions. In scenarios involving rapid posture changes and swift movements, it is a challenging task for a motion predictor to swiftly acquire and incorporate motion information and make prompt adjustments to motion prediction. Therefore, the efficacy of modeling motion information based on past trajectories is considerably diminished. During complex movements, the motion direction of an object can shift significantly, making motion direction information a crucial factor in predicting future object motion. We propose a novel loss function called Momentum Correction Loss (MCL), which is employed as a regularizer for the primary motion prediction task. During training, the motion predictor is guided by a loss function that encourages the predicted motion directions to align with the actual motion direction. In MOT tasks, the position of an object is typically represented by a bounding box. In scenarios where the posture of some objects suddenly changes\citep{sun2022dancetrack}, we consider not only the motion directions of the center point of the objects but also the motion directions of their four corners. By incorporating additional motion direction information, the model predictor can rapidly adapt to motion variations and predict future motion more accurately.
The contributions of this study can be summarized as follows:
\begin{enumerate}
\item{We propose an enhanced temporal motion predictor that integrates a Temporal Convolutional Network (TCN) and a Temporal Transformer model. It can effectively capture and comprehend objects' motion patterns to improve object tracking performance.}
\item{We introduce a novel momentum correction loss function that provides the motion predictor with additional information about the motion direction of objects by correcting the motion direction of objects during learning.}
\item{We demonstrate that our proposed method achieves competitive performance on challenging datasets, such as DanceTrack \citep{sun2022dancetrack} and SportsMOT \citep{cui2023sportsmot}, where non-linear movements, diverse poses, and severe occlusion are present. Furthermore, our method achieves comparable results on MOT17\citep{milan2016mot16}.}
\end{enumerate} 

\begin{figure*}[t]
\centering
\includegraphics[width=1\textwidth]{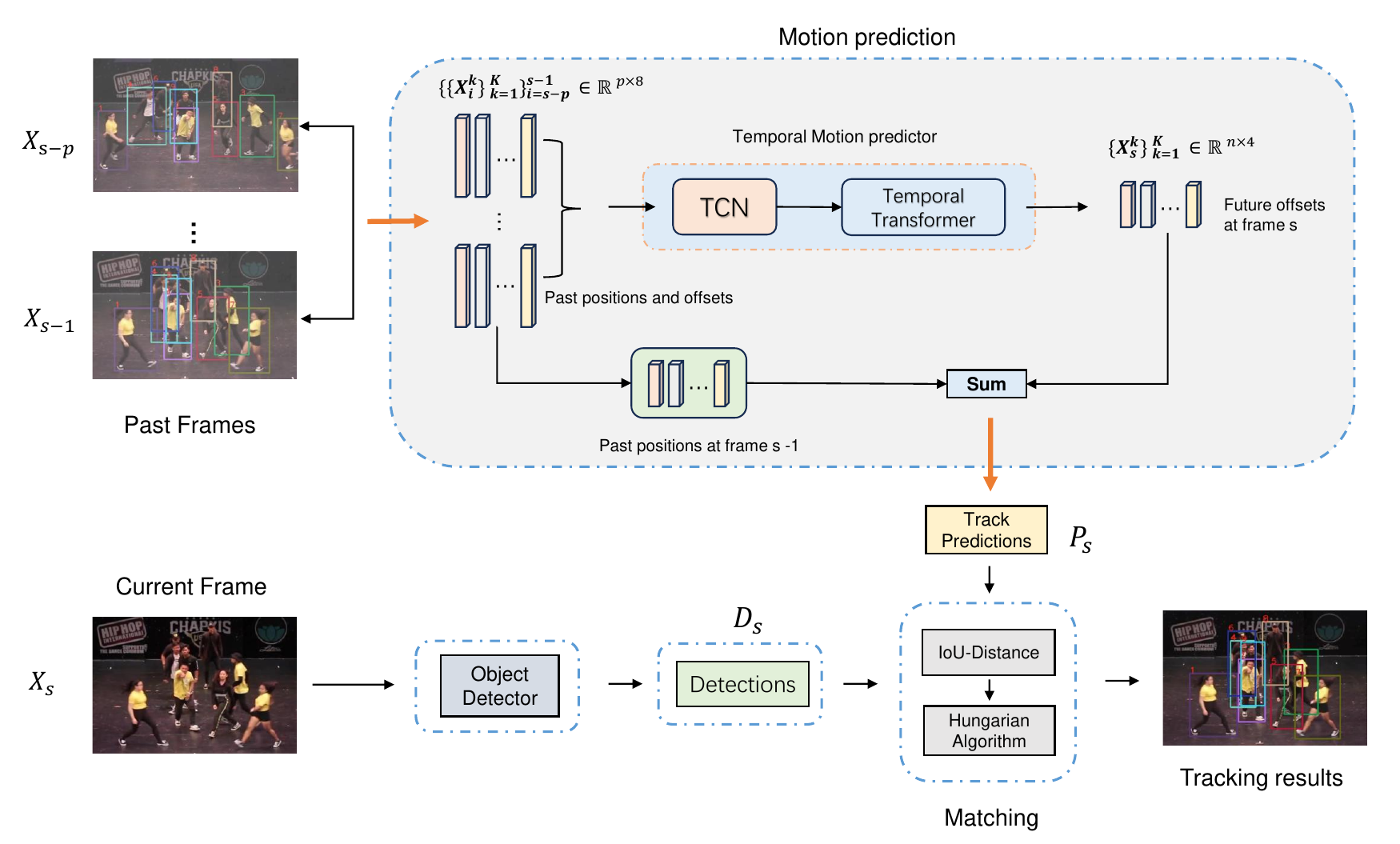}
\caption{The pipeline of ETTrack. The historical trajectory of length $p$ is fed into our motion predictor, which predicts the track predictions $P_s$ at the current moment $s$. By deploying an object detector, detections $D_s$ are obtained. With the track predictions and detections, data association can be accomplished by IoU Distance and Hungarian matching algorithm. Different colors represent different identities in the tracking results}\label{pipeline}
\end{figure*}

\section{Related Works}\label{sec2}
\subsection{Multi-Object Tracking}
Multi-Object Tracking methods can be categorized into two types: tracking-by-detection and joint-detection-tracking. Tracking-by-detection methods, such as SORT\citep{bewley2016simple} and DeepSORT\citep{wojke2017simple}, first detect objects and then associate them using appearance and motion information. These methods have long been the dominant paradigm in MOT. Alternatively, joint-detection-tracking methods, such as JDE\citep{wang2020towards} and FairMOT\citep{zhang2021fairmot}, incorporate the detection and ReID model for joint training, offering comparable performance with low computational costs. However, joint-detection-tracking methods may face reduced efficiency owing to conflicts between the detection and tracking optimization goals in their unified network. Additionally, ByteTrack\citep{zhang2022bytetrack} utilizes a simple yet effective data association method BYTE to significantly enhance the tracking accuracy and robustness. Recently, BotSORT\citep{aharon2022bot} with a stronger ReID model outperforms ByteTrack by leveraging both motion and appearance information. Thus, Tracking-by-detection methods demonstrate that a robust detector combined with a simple association approach can attain good tracking results. Therefore, we chose to follow the ByteTrack algorithm replacing the Kalman Filter\citep{kalman1960contributions} with a deep-learning-based motion model.

\subsection{Motion Model}

Several mainstream MOT algorithms\citep{bewley2016simple,wojke2017simple,cao2023observation,choi2015near} use motion models. Typically, SORT-series trackers\citep{bewley2016simple,zhang2022bytetrack,cao2023observation} utilize the Bayesian estimation\citep{lehmann2006theory} as a motion model to predict the subsequent state by maximizing posterior estimation. For instance, SORT\citep{bewley2016simple} utilizes the classic Kalman Filter\citep{kalman1960contributions}, assuming linear motion for object estimation, and the Hungarian matching algorithm\citep{kuhn1955hungarian} to match predictions and detections. OC\_SORT\citep{cao2023observation} enhances robustness in handling occlusions by prioritizing object observations rather than linear state estimations, but it still suffers from long-term occlusion and struggles in recovering lost objects undergoing non-linear motion. However, as has been emphasized, Kalman Filter based methods presuppose a constant motion, which does not accurately describe the change in object positions when undergoing complex interactions within a scene. Therefore, some MOT methods introduce deep-learning-based motion models\citep{milan2017online,kesa2021joint} for non-linear motion modeling. For example, \citep{milan2017online} presents a novel tracker based on recurrent neural networks (RNNs) for online MOT. \citep{kesa2021joint}proposes a joint learning architecture for improved MOT and trajectory forecasting by leveraging the capabilities of RNNs and adding additional appearance information, thereby surpassing the limitations imposed by using a traditional Kalman Filter. The DEFT algorithm\citep{chaabane2021deft} uses  an LSTM to capture the motion constraints of objects. \citep{xiao2023motiontrack} proposes a Transformer-based motion model to capture long-range dependencies for modeling motions. However, these current approaches lack the capacity to model more intricate temporal dependencies and integrate reliable additional information, resulting in inadequate motion prediction capabilities in complex and non-linear scenarios. The proposed method addresses these limitations and enhances the predictive capabilities.

\subsection{Transformer-based Methods}

Since the Transformer\citep{Vaswani_Shazeer_Parmar_Uszkoreit_Jones_Gomez_Kaiser_Polosukhin_2017} has become popular in computer vision, many methods\citep{meinhardt2022trackformer,zeng2022motr,sun2012transtrack,chu2023transmot,zhang2023motrv2} for the MOT task have been proposed to leverage its powerful attention mechanism to extract deep representations from both visual information and object trajectories. For example, TrackFormer\citep{meinhardt2022trackformer} and MOTR\citep{zeng2022motr} extend from Deformable DETR\citep{zhu2020deformable}. They utilize both track queries and standard detection queries to predict object bounding boxes and associate the same objects in subsequent frames. TransTrack\citep{sun2012transtrack} employs only Transformers as its feature extractor and propagates track queries once to obtain the position of objects in the subsequent frame. TransMOT\citep{chu2023transmot} uses convolutional neural networks (CNNs) as a detector to extract features and employs spatial-temporal transformers to learn an affinity matrix. Recently, MOTRv2\citep{zhang2023motrv2} combines a separate detector with MOTR\citep{zeng2022motr} to address the conflict between the detection and association. However, Transformer-based methods require extensive training time and computational resources, which prevents them from achieving real-time capability. In contrast, the proposed method utilizes the powerful temporal dependency modeling capabilities of a Transformer to model the movement of objects. In addition, the ETTrack method relies solely on trajectory data as input, which significantly decreases the computation time required for motion model inference during the runtime.

\section{Method}\label{sec3}
In this work, we propose an enhanced temporal motion predictor that effectively utilizes motion cues to track objects with complex motion patterns. Our primary objective is to achieve precise estimates of non-linear uncertainties by integrating a Temporal Transformer with a Temporal Convolutional Network (TCN) that surpasses the performance of some deep-learning-based motion models. In addition, we propose a momentum correction loss function to enhance the motion predictor by using motion direction information. Section \ref{subsubsec1} and Section \ref{subsubsec2}  describe the Temporal Transformer and the Temporal Convolutional Network (TCN) respectively, while Section \ref{subsubsec3} introduces the concept of momentum correction loss function.

\subsection{Problem Formulation}\label{subsec1}
The trajectory of an individual object $i$ contains a sequence of bounding boxes $ \mathbf{B}=\left\{b_{t_{1}}, b_{t_{2}}, \ldots, b_{t_{N}}\right\}$, where $t$ stands for the timestamp, and $N$ is the total number of frames. The bounding boxes are represented as b=$(x,y,w,h)$. The aim of MOT is to assign a unique identifier to all the frame-wise bounding boxes. This assignment aims to establish a comprehensive association between all the bounding boxes. 

Our goal is to create a motion predictor that predicts the locations of objects. When the historical trajectory length of 
objects is set to $p$, the historical trajectory of objects can be denoted as a sequence $\mathbf{X} \equiv \{\{X_i^k\}_{k=1}^K\}_{i=s-p}^{s-1}\equiv\{\{X_{s-p}^k\},\{X_{s-
p+1}^k\},...,\{X_{s-1}^k\}\} \in \mathbb{R}^{p\times8}$, where $K$ is the total number of objects across all frames and $s$ is the frame index.  The object $k$ at the moment $s-1$ is denoted by $\mathbf{X}_{s-1}^k=\left(x_{s-1}^k, y_{s-1}^k, w_{s-1}^k, h_{s-1}^k, \Delta x_{s-1}^k, \Delta y_{s-1}^k, \Delta w_{s-1}^k, \notag\right.\\ \left. \Delta h_{s-1}^k\right)$, where $(x_{s-1}^k,y_{s-1}^k)$ are the center coordinate of the corresponding bounding box, $(w_{s-1}^k,h_{s-1}^k)$ stand for the width and height of the bounding box, respectively, and $\mathbf{V}_{s-1}^k=\left(\Delta x_{s-1}^k, \Delta y_{s-1}^k, \Delta w_{s-1}^k, \Delta h_{s-1}^k\right)$ is the velocity of the object. These historical trajectories are fed into the motion predictor to predict the velocity $\mathbf{V}_s^k=\left(\Delta x_s^k, \Delta y_s^k, \Delta w_s^k, \Delta h_s^k\right)$  at the current moment $s$. 

Predicting offsets (i.e., position changes) between successive frames offers two advantages for motion prediction. First, predicting position changes simplifies the overall prediction task because it involves analyzing relative movements, which typically exhibit less variation and more predictable patterns than the absolute positions. By focusing on relative movements, the complexity of the prediction process is reduced. Second, our motion predictor becomes less sensitive to specific starting points or trajectory shapes. This advantage enables the motion predictor to effectively capture the underlying movement dynamics, thereby improving its ability to generalize across different scenarios and unseen data. 

To obtain the predicted position at the current moment $s$, the motion predictor adds bounding boxes $\left\{X_{s-
1}^k\right\}_{k=1}^K$ from the last frame to the predicted velocities to generate the predicted current bounding boxes
$\mathbf{X}_ {s}^ {k} =\left\{ x_ {s}^ {k} , y_ {s}^ {k} , w_ {s}^ {k} , h_ {s}^ {k} \right\}=\left\{ x_ {s-1}^ {k} + \Delta x_ 
{s}^ {k} , y_ {s-1}^ {k} + \Delta y_ {s}^ {k} , w_ {s-1}^ {k} + \Delta w_ {s}^ {k} , h_ {s-1}^ {k} + \notag\right.\\ \left. \Delta h_ {s}^ 
{k}\right\}$. The overall framework of the motion predictor is illustrated in Fig. \ref{pipeline}. 

\subsection{Motion Predictor}\label{subsec2}

The MOT task requires identification of the spatial and temporal locations of objects, specifically their trajectories. It has been demonstrated that the Temporal Transformer can capture global temporal dependencies and comprehend the overall context of motion sequences, offering a comprehensive perspective of long-range motion interactions and patterns. Moreover, the Temporal Convolutional Network (TCN)\citep{bai2018empirical} has proven effective in identifying complex local temporal dependencies in motion sequences, resulting in precise analysis of short-term motion patterns. Thus, the integration of the Temporal Transformer and TCN allows the model to provide a comprehensive understanding of the motion patterns.

\begin{figure}[t]
\centering
\includegraphics[width=0.4\textwidth]{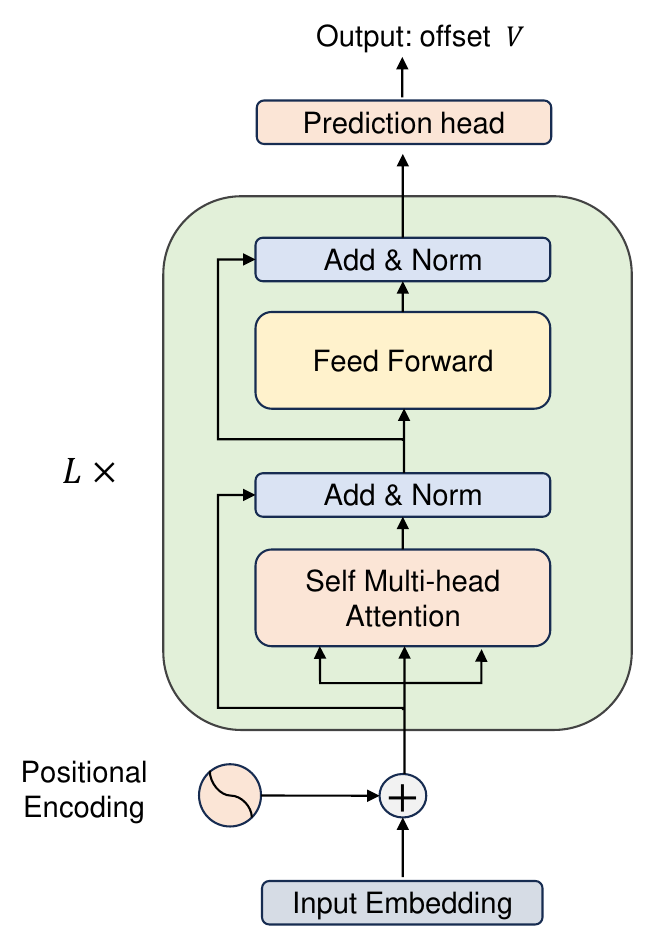}
\caption{The network structure of Temporal Transformer in the motion predictor}\label{transformer}
\end{figure}

\subsubsection{Temporal Transformer}\label{subsubsec1}
The Temporal Transformer is designed to effectively capture the long-term historical motion patterns of individual objects. This is achieved by utilizing a standard transformer encoder, which comprises a multi-head self-attention (MHSA) mechanism. The MHSA enables the encoder to consider various aspects of the trajectory sequence and identify the most critical features for predicting the future positions of objects. The structure of the Temporal Transformer is shown in Fig. \ref{transformer}.

In the Temporal Transformer, the input sequence of token T is the output of the Temporal Convolutional 
Network. The self-attention of the temporal transformer can learn the query matrix $Q=f_Q(T)$, 
key matrix $K=f_K(T)$, and value matrix $V=f_V(T)$. Self-attention of a single head is 
calculated as:\begin{equation}\label{1}Attention(Q,K,V)=\frac{softmax(QK^T)}{\sqrt{d_k}}V,\end{equation}where  
$:\frac1{\sqrt{d_{k}}}$accounts for the numerical stability of self-attention. In Eq.\ref{1}, $softmax$ is the distribution function, which depends on the properties of the model. Utilizing multiple matrices for attention recurrence allows for a more effective handling of complex temporal dependencies. This is achieved through the implementation of multi-head attention, which involves embedding the outputs of multiple self-attention mechanisms. This method enables the model to simultaneously consider information from various subspaces of representation at different positions, thereby enhancing its ability to process and understand complex information. With $n$ heads, the multi-head attention can be represented as:\begin{equation} \label{2}Multi\,\,Attention=f_h(\mathrm{Attention}
(Q_i,K_i,V_i)_{i=1}^n),\end{equation}
\begin{figure}[t]
\centering
\includegraphics[width=0.35\textwidth]{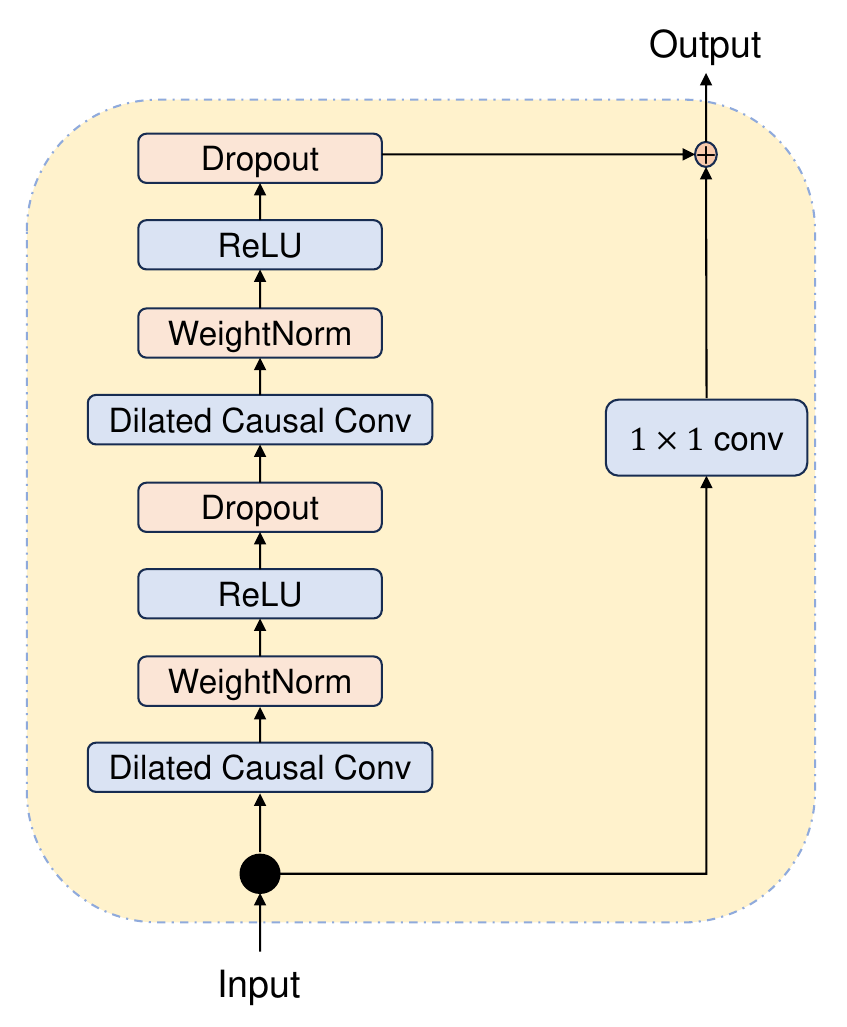}
\caption{Structure of the TCN}\label{tcn}
\end{figure}
\begin{figure}[t]
\centering
\includegraphics[width=0.45\textwidth]{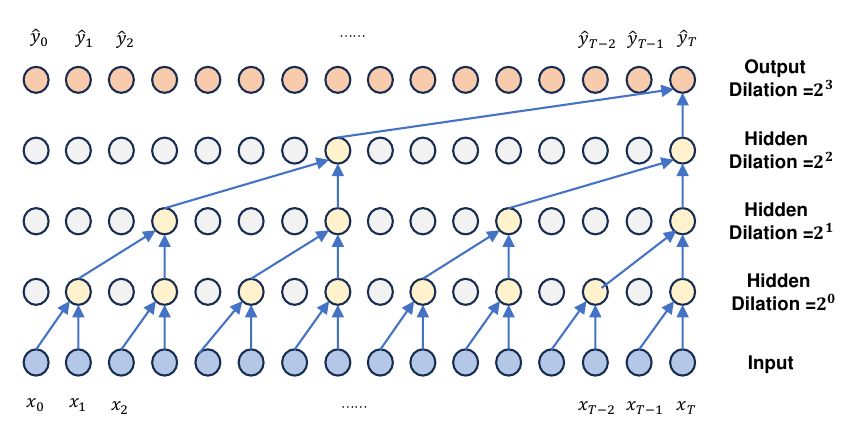}
\caption{visualization of a stack of dilated causal convolutions ($K$ = 7, Dilation = [$\mathbf{2}^\mathbf{0}$,$\mathbf{2}^\mathbf{1}$,$\mathbf{2}^\mathbf{2}$,$\mathbf{2}^\mathbf{3}$])}\label{dilation}
\end{figure}
where $f_{h}$ is a fully connected feed-forward network. A 
positional encoding method is implemented to provide the Transformer encoder with positional 
information and to enable the attention layer to perform multi-head self-attention on the output of 
the TCN. Finally, the output of $n$ heads is cascaded and fed into  $f_{h}$ to obtain the Transformer 
output.
The temporal Transformer is an important implementation of the Transformer model for modeling object motion by processing sequential data. In our experiments, we demonstrate that the temporal Transformer can efficiently capture the temporal interdependencies within the input trajectory sequence.

\subsubsection{Temporal Convolutional Network} \label{subsubsec2}
The Temporal Convolutional Network (TCN) is an innovative structure optimized for sequential data processing that effectively addresses temporal dependencies using a deep learning approach. A standard TCN is composed of multiple TCN blocks, each designed to capture temporal patterns and ensure robust feature representation. Details of the TCN are shown in Fig. \ref{tcn}. Each TCN block consists of successive layers, beginning with a pair of causal convolutions. These convolutions ensure that the model's predictions are solely dependent on past and present information rather than future data. Although causal convolutions have been shown to be effective in modeling short-term dependencies, they have limitations in modeling long-term dependencies because of their limited receptive fields. To enhance the capacity of the TCN to capture longer temporal interdependencies, we can either increase the network depth or the filter size. However, this results in higher computational complexity. Unlike causal convolution, which can only broaden its receptive field by increasing its kernel size or adding more layers, dilated causal convolution employs dilation factors, represented by $dil\in F$, to enlarge its receptive field. The dilation factor, denoted by $dil$, can be exponentially increased, as shown in Fig. \ref{dilation}. Using dilated causal convolution, we can calculate the feature map $f$ as follows:
\begin{equation}\label{3}f(t)=(I*_{dil}W)(t)\sum_{i=0}^{K-1}w(i)I(t-idil),\end{equation} where $I$ is the input, $W$ represents the filter, and $K$ is the filter size. By stacking these convolutions, a TCN can expand its receptive field and effectively capture longer dependencies. Following dilated causal convolution, two layers of non-linearity are introduced using the ReLU activation function. This non-linearity is crucial for the model to capture complex patterns and relationships in data. Weight normalization is added to the one-dimensional convolutions to improve the training stability and speed. In addition, a dropout block is added after each activation function. Finally, a residual connection is integrated into the layer to enhance the predictive performance of the model. Residual blocks \citep{zagoruyko2016wide} equipped with identity mapping can be denoted as: \begin{equation}\label{4}T_{j+1}=T_j+\mathcal{R}(T_j,W_j),\end{equation} Where $T_{j}$ and $T_{j+1}$ represent the input and output of the ($j$+1)th TCN block, respectively, $W_{j}$ is the trainable parameter matrix of the residual blocks, and $\mathcal{R}(\cdot)$ denotes a residual function.

Although the TCN can be designed to enhance the capacity to model long-term dependencies using dilated causal convolution, it is equally optimized for modeling short-term dependencies, allowing it to accurately capture local patterns and dynamics in sequence data. The TCN effectively mitigates the limitations of the Temporal Transformer in modeling short-term dependencies, when the tracking task is particularly dependent on localized features in the short term. Moreover, compared with the Temporal Transformer, which requires complex self-attention computation, the TCN typically has a simpler model structure. This not only alleviates computational demands, but also allows TCN models to accomplish the training and inference phases more rapidly in real-world scenarios.

\subsubsection{Momentum Correction Loss (MCL)}\label{subsubsec3}

\begin{figure}[t]
\centering
\includegraphics[width=0.47\textwidth]{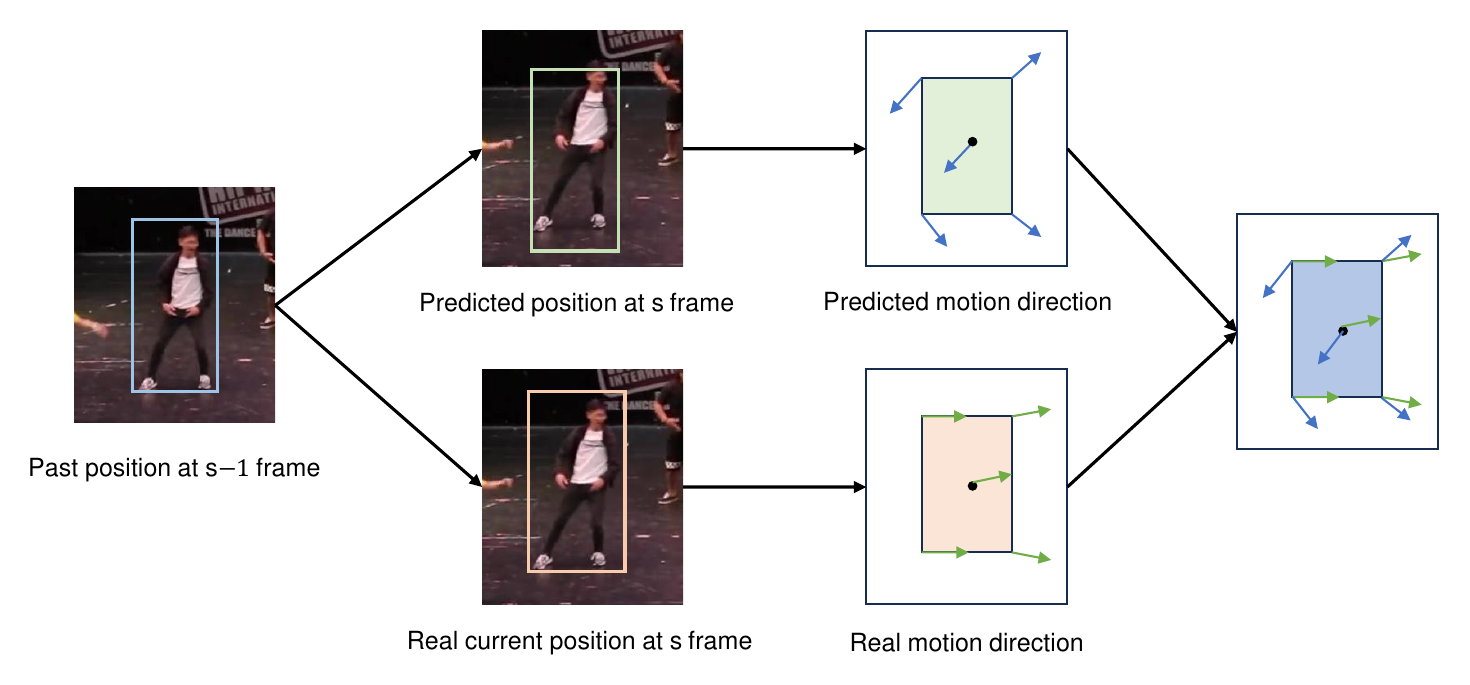}
\caption{Illustration of momentum correction loss. We obtain the predicted and actual motion directions, which are represented by the \textcolor{blue}{blue} and \textcolor{green}{green} arrows, respectively. Subsequently, we calculate the angular difference between the predicted and actual motion directions to obtain the momentum correction loss}\label{direction}
\end{figure}

To enhance the motion predictor model and obtain more reliable future predictions, we incorporate contextual information, such as motion direction. We propose a novel loss called momentum correction loss, which serves as a regularizer for the motion prediction loss. By integrating both the trajectory prediction loss and momentum correction loss, we can effectively train the motion predictor model. The proposed momentum correction loss is shown in Fig. \ref{direction}. Given two points $(x_1,y_1)$ and $(x_2,y_2)$, the motion direction is denoted by:\begin{equation}\label{5}\theta=\arctan\left(\frac{y_1-y_2}{x_1-x_2}\right).\end{equation} In MOT, the position of an object is typically represented by a bounding box. To adjust to sudden pose changes and swift movements, the predicted object motion direction can be represented as $\theta_p=\{\theta_c,\theta_{lt},\theta_{rt},\theta_{lb},\theta_{rb}\},$ where $\{\theta_c\}$ is the motion direction at the center point and $\{\theta_{lt},\theta_{rt},\theta_{lb}{,\theta}_{rb}\}$ represent the motion direction at the four corners of the object. The real object motion direction can also be denoted as $\theta_t=\{\theta_c^{\prime},\theta_{lt}^{\prime},\theta_{rt}^{\prime},\theta_{lb}^{\prime},\theta_{rb}^{\prime}\}$. In general, the direction of motion of the center point of an object represents the overall direction of motion. In scenarios where the posture of some objects suddenly changes\citep{sun2022dancetrack}, it is necessary to consider not only the direction of motion of the center point of an object but also the direction of motion of the four corners of the bounding box. As illustrated in Fig. \ref{direction}, we can capture more pose changes using the four corners of the bounding box. Thus, the momentum correction loss can be calculated as: \begin{equation}\label{6}\mathcal{L}_{MCL}=\frac15\sum_i|\theta_i-\theta_i^{\prime}|, i\in(c,lt,rt,lb,rb).\end{equation}
\subsection{Training and Inference}

\textbf{Training.} In the training phase, we train the motion predictor by utilizing the historical positions of n frames and predict the object positions of the subsequent frames. We use the L1 loss function as the prediction loss to supervise the training process and improve the capacity to address outliers. Specifically, given the predicted offset $\mathbf{\widehat{V}}=\left\{\widehat{v}_x,\widehat{v}_y,\widehat{v}_w,\widehat{v}_h\right\}$, and the corresponding attributes of the ground truth $V$, the prediction loss $\mathcal{L}_{pred}$ is obtained by: \begin{equation}\label{7}\mathcal{L}_{pred}(\mathbf{\hat{V}},\mathbf{V})=\sum_i|\hat{v}_i-v_i|\text{,}i\in(x,y,w,h).\end{equation}
The final loss function sums both the motion prediction loss $\mathcal{L}_{pred}$ and momentum correction loss $\mathcal{L}_{MCL}$:\begin{equation}\label{8}\mathcal{L}_{final}=\mathcal{L}_{pred}+\beta\mathcal{L}_{MCL},\end{equation} where $\beta$ is a crucial hyper-parameter that determines the extent to which the momentum correction loss function is influential. It is worth noting that additional direction information is required only during the training phase. By contrast, during the prediction phase, the model relies solely on the observed trajectory to predict the future trajectory.

\section{Expreriments}
\subsection{Datasets and Evaluation Metrics}
\begin{algorithm}
\small
\caption{Pseudo-code of ETTrack}\label{algo1}
\renewcommand{\algorithmicrequire}{\textbf{Input:}}
\renewcommand{\algorithmicensure}{\textbf{Output:}}
\begin{algorithmic}[1]
\Require A video sequence $V$; object detector \textbf{D}; motion predictor \textbf{T}; detection score threshold $\tau $ 
\Ensure Tracks $\mathcal{T}$of the video

\State initialization: $\mathcal{T}\Leftarrow\emptyset $
\For{$frame$ $f$ $in$ $V$}
         \State    /*Detection      */
          \State $D_f \Leftarrow \textbf{D}(f)$ 
          \State $D_{high} \Leftarrow \emptyset$
          \State $D_{low} \Leftarrow \emptyset$ 
         
          \For{$d$  $in$ $D_f$}
            \If{ $d.score \geq \tau $}
                 \State $D_{high}\Leftarrow{D}_{high}\cup\{d\}$
            \Else
                  \State $D_{low}\Leftarrow{D}_{low}\cup\{d\}$
            \EndIf
          \EndFor
         \State    /*motion predictor    */
         \For{ $trks$ in $\mathcal{T}$}
              \State    $trks \Leftarrow $  \textbf{T}($trks$)
          \EndFor
\State    /*first association      */
\State   Associate $\mathcal{T}$ and $D_{high}$ using IOU
\State $D_{remian}\Leftarrow$\, remaining detected boxes from ${D}_{high}$
\State $\mathcal{T}_{remian}\Leftarrow $\, remaining predicted boxes from $\mathcal{T}$
\State    /*second association      */
\State   Associate $\mathcal{T}_{remian}$ and $D_{low}$ using IOU
\State $\mathcal{T}_{re-remian}\Leftarrow $\, remaining predicted boxes from $\mathcal{T}_{remian}$
\State    /*delete unmatched tracks      */
\State  $\mathcal{T}\Leftarrow \mathcal{T}\backslash \mathcal{T}_{re-remian}$
\State    /*initialize new tracks      */
          \For{ $d$ in $D_{remian}$}
          \State $\mathcal{T}\Leftarrow \mathcal{T}\cup\{d\}$
          \EndFor

\EndFor
\State Return $\mathcal{T}$
\end{algorithmic}
\end{algorithm}

\textbf{Inference.} Our motion predictor is applied to the ByteTrack \citep{zhang2022bytetrack} platform, which introduces a two-step association algorithm that utilizes object detection thresholds to track every detection box using tracklets and recovers occluded objects based on similarities. We use the motion predictor to output the prediction boxes \textbf{$P_s$} in the current frame. By utilizing a YOLOX\citep{ge2021yolox} detector, the detection boxes \textbf{$D_s$} are obtained in the current frame. ByteTrack's association algorithm employs the Hungarian algorithm to assign detections to tracklets based on the Intersection-over-Union (IoU) between \textbf{$P_s$} and \textbf{$D_s$}. These successfully associated detection boxes are incorporated into the historical trajectories for updates, and the unassigned detections are then initialized as new trajectories. For inactive trajectories, our motion predictor continues to generate prediction boxes, that are subsequently appended to trajectories to facilitate model inference. When detections are reconnected to these inactive trajectories, the inactive trajectories may also be retracked. If the inactive time exceeds a given threshold, these preserved prediction boxes are deleted. The pseudo-code of ETTrack is shown in Algorithm \ref{algo1}. 

\begin{table*}[h]
%\small
\footnotesize
\renewcommand\arraystretch{1.3}
\caption{Comparison of our method with start-of-the-art MOT algorithms on the DanceTrack test sets}\label{dance}%
\begin{threeparttable}
\begin{tabular*}{\textwidth}{@{\extracolsep\fill}lccccc}
%%\begin{tabular}{@{}lccccc@{}}
\toprule
Tracker&HOTA$\uparrow$ &DetA $\uparrow$  &AssA$\uparrow$ &MOTA$\uparrow$ &IDF1$\uparrow$\\
\midrule%第二道横线 
CenterTrack\citep{zhou2020tracking}	&48.1	&86.8	&78.1	&22.6	&35.7 \\
FairMOT\citep{zhang2021fairmot} &39.7&	82.2	&66.7	&23.8&	40.8 \\
QDTrack\citep{fischer2023qdtrack} &45.7&	83&	72.1&	29.2&	44.8 \\
TransTrack\citep{sun2012transtrack} &45.5&	88.4&	75.9&	27.5&	45.2 \\
TraDes\citep{wu2021track} &43.3	&86.2	&74.5	&25.4	&41.2 \\
GTR\citep{zhou2022global} &48	&84.7	&72.5	&31.9	&50.3 \\
MORT\citep{zeng2022motr} &54.2	&79.7	&73.5	&\textbf{40.2}	&51.5 \\
\midrule%第二道横线
MotionTrack\citep{xiao2023motiontrack} &52.9  &91.3 & 80.9  &34.7  &53.8  \\
SORT\citep{bewley2016simple}	&47.9	&91.8	&72	&31.2	&50.8 \\
DeepSORT\citep{wojke2017simple}	&45.6	&87.8	&71	&29.7	&47.9 \\
ByteTrack\citep{zhang2022bytetrack} &47.3	&89.5	&71.6	&31.4	&52.5 \\
OC\_SORT\citep{cao2023observation} &55.1	&89.4	&80.3	&38	&54.2  \\
ETTrack(ours) &\textbf{56.4}	&\textbf{92.2}	&\textbf{81.7}	&39.1	&\textbf{57.5}  \\
\botrule
\end{tabular*}
\footnotesize{The best results of methods are marked in bold} \\
\footnotesize{Methods in bottom block use the same YOLOX detector}
\end{threeparttable}
\end{table*}

\begin{table*}[h]
%\small
\renewcommand\arraystretch{1.3}
\footnotesize
\caption{Comparison of our method with start-of-the-art MOT algorithms on the SportsMOT test sets}\label{sport}%标题
\begin{threeparttable}
\begin{tabular*}{\textwidth}{@{\extracolsep\fill}lcccccccc}
\toprule%
Tracker&	HOTA$\uparrow$&	IDF1$\uparrow$&	AssA$\uparrow$&	MOTA$\uparrow$&	DetA$\uparrow$&	LocA$\uparrow$&	IDs$\downarrow$&	Frag$\downarrow$\\
 
\midrule%第二道横线 
FairMOT\citep{zhang2021fairmot}&	49.3&	53.5&	34.7&		86.4&	70.2&	83.9&	9928&	21673\\
QDTrack\citep{fischer2023qdtrack}&	60.4&	62.3&	47.2&		90.1&	77.5&	88&	6377&	11850\\
CenterTrack\citep{zhou2020tracking}&	62.7&	60&	48&		90.8&	82.1&	90.8&	10481&	5750\\
TransTrack\citep{sun2012transtrack}&	68.9&	71.5&	57.5&		92.6&	82.7&	91&	4492&	9994\\
\midrule%第二道横线 
BoT-SORT\citep{aharon2022bot}&	68.7&	70&	55.9&		94.5&	84.4&	90.5&	5729&	5349\\
ByteTrack\citep{zhang2022bytetrack}&	62.8&	69.8&	51.2&		94.1&	77.1&	85.6&	3267&	4499\\
OC\_SORT\citep{cao2023observation}&	71.9&	72.2&	59.8&		94.5&	86.4&	92.4&	3093&	3474\\
ETTrack&	72.2&	72.5&		60.1&	94.9&	86.9&	92.5&	4075&	5279\\
\midrule%第二道横线 
$\ast$ByteTrack\citep{zhang2022bytetrack}&	64.1&	71.4&	52.3&		95.9&	78.5&	85.7&	3089&	4216\\
$\ast$MixSort\_Byte\citep{cui2023sportsmot}&		65.7&	74.1&	54.8&		96.2&	78.8&	85.7&	\textbf{2472}&	4009\\
$\ast$OC\_SORT\citep{cao2023observation}&	73.7&	74&	61.5&		96.5&	88.5&	92.7&	2728&	\textbf{3144}\\
$\ast$MixSort\_OC\citep{cui2023sportsmot}&		74.1&	74.4&	62&		96.5&	88.5&	92.7&	2781&	3199\\
$\ast$ETTrack &		\textbf{74.3}&	\textbf{74.5}&		\textbf{62.1}&	\textbf{96.8}&	\textbf{88.8}&	\textbf{92.8}&	3862&	4298\\
\botrule
\end{tabular*}
\footnotesize{The best results of methods are marked in bold} \\
\footnotesize{Methods in middle and bottom block use the same YOLOX detector}\\
\footnotesize{Methods with $\ast$ show that their YOLOX detectors are trained on the SportsMOT train and validation sets}
\end{threeparttable}
\end{table*}

\textbf{Datasets.} To conduct a comprehensive evaluation of our method, we performed experiments on various MOT benchmarks including DanceTrack \citep{sun2022dancetrack}, SportsMOT \citep{cui2023sportsmot} and MOT17 \citep{milan2016mot16}. MOT17 is a widely used foundational benchmark in MOT, in which the motion of pedestrians is mostly linear. In contrast, the SportsMOT dataset captures the intricate motions and similar appearances of athletes in sports scenes, incorporating videos from high-profile events, such as the Olympic Games and NBA, thereby demanding high tracking precision. The DanceTrack dataset presents a particularly complex challenge in terms of object tracking. This is because it consists of objects that look very similar to each other, frequently become occluded from view, and exhibit unpredictable movement patterns. Consequently, it is challenging for any tracking algorithm to conclusively demonstrate its ability to  handle complex scenarios effectively. Our aim is to propose a motion predictor that can effectively improve the tracking performance in challenging situations, particularly when the Kalman Filter fails under diverse scenarios. SportsMOT and DanceTrack are ideal datasets for evaluating the tracking performance.

\textbf{Metrics.} To evaluate our algorithm, we adapt HOTA \citep{luiten2021hota} as the primary metric. HOTA combines several sub-metrics and provides a balanced view by considering both detection and association accuracy. In addition to HOTA, we also used other CLEAR metrics \citep{bernardin2008evaluating}, such as MOTA, FP, FN, IDs, \textit{etc.}, and IDF1\citep{ristani2016performance}. MOTA is computed based on FP, FN and IDs and is affected by the detection performance. IDF1 evaluates identity preservation ability and is used to measure association performance. These metrics are widely used to effectively evaluate the algorithm performance.

 %经典三线表
\begin{table*}[h]
%\small
\footnotesize
\renewcommand\arraystretch{1.3}
\caption{Tracking performance of investigated algorithms on MOT17 dataset}%标题
\label{mot17}
\begin{threeparttable}
\begin{tabular*}{\textwidth}{@{\extracolsep\fill}lcccccccccc}
%%\begin{tabular}{l|cccccccccc}%四个c代表该表一共四列，内容全部居中
\toprule%第一道横线
Tracker&HOTA$\uparrow$ &MOTA $\uparrow$ &IDF1$\uparrow$ &FP$\downarrow$ &FN $\downarrow$ &IDs$\downarrow$ $\downarrow$ &AssA$\uparrow$ &AssR$\uparrow$ \\

\midrule%第二道横线 
FairMOT\citep{zhang2021fairmot} &59.3  &73.7  &72.3  &27500  &117000  &3,303    &58.0  &63.6 \\
QDTrack\citep{fischer2023qdtrack} &53.9  &68.7  &66.3  &26600  &147000  &3,378   &52.7  &57.2 \\
TransTrack\citep{sun2012transtrack} &54.1  &75.2  &63.5  &50200  &864000  &3,603   &47.9  &57.1 \\
TransCenter\citep{9964258}  &54.5  &73.2  &62.2  &23100  &124000  &4,614    &49.7  &54.2 \\
MORT\citep{zeng2022motr} &57.2  &71.9  &68.4  &21100  &136000  &2,115    &55.8  &59.2 \\
TransMOT\citep{chu2023transmot} &61.7 &76.7 &75.1 &36200 &93200 &2,346  &59.9 &66.5 \\
GTR\citep{zhou2022global}  &59.1  &75.3  &71.5  &26800  &110000  &2,859   &61.6 &- \\
\midrule%第二道横线
ByteTrack\citep{zhang2022bytetrack} &63.1  &\textbf{80.3}  &77.3  &25500  &\textbf{83700}  &2,196  &62.0  &\textbf{68.2} \\
StrongSORT\citep{du2023strongsort} &\textbf{63.5}  &78.5  &\textbf{78.3}  &-  &-  &\textbf{1,446}    &63.7  &- \\
OC\_SORT\citep{cao2023observation} &63.2  &78.0  &77.5  &\textbf{15100}  &108000  &1,950    &\textbf{63.2}  &67.5 \\
ETTrack&	61.9&	79.0&	75.9&		23100&	93300 &2,118&		60.5&	67.0\\

\bottomrule%第四道横线
\end{tabular*}
\footnotesize{The best results of methods are marked in bold} \\
\footnotesize{Methods in bottom block use the same YOLOX detector}
\end{threeparttable}
\end{table*}

\subsection{Implementation Details}

We train the motion predictor solely on the corresponding tracking datasets, without integrating any external samples. In the experiments, we use the publicly available YOLOX\citep{ge2021yolox} detector weights developed by ByteTrack\citep{zhang2022bytetrack} for a fair comparison. For the motion predictor, a TCN is employed to encode the past trajectory into a 32-dimensional vector. The TCN comprises 4 TCN blocks. Moreover, a dropout ratio of 0.1 is implemented during data processing. It is worth noting that all the Temporal Transformer layers accept inputs with a feature size of 32. The Temporal Transformer includes 6 layers, and the multi-head self-attention uses 8 heads. We optimize the network using the Adam algorithm \citep{Kingma_Ba_2014} algorithm with a learning rate of 0.0015 and a batch size of 16 for 50 epochs. The maximum historical trajectory length $p$ is set to 10. On the training datasets, we conduct hyper-parameter optimization of $\beta$. We achieve best tracking results using $\beta$ = 0.3 on the DanceTrack validation sets. All the experiments are conducted using a GeForce RTX 3090 GPU.

\subsection{Benchmark Evaluation}

Here, we present benchmark results for multiple datasets, such as DanceTrack, SportsMOT, and MOT17. All these methods use the same detection results. 

\textbf{DanceTrack.} To demonstrate the performance of ETTrack with non-
linear object motion and diverse scenarios, the results for the 
DanceTrack dataset are presented in Table \ref{dance}. The performance 
of our method is tested on the DanceTrack test sets. The results 
demonstrate that ETTrack performs competitively compared with the other
methods. Specifically, ETTrack achieved 56.4$\%$ HOTA, 92.2$\%$ DetA 
and 57.5$\%$ IDF1, which is better than the 
OC\_SORT method with an enhanced Kalman 
Filter and recovery strategy. It is noteworthy that a significant gain 
of 3.4$\%$ is observed in the IDF1 score, which assesses the 
performance of association accuracy. These experimental results provide
convincing evidence that our method outperforms SORT-like algorithms 
\citep{bewley2016simple,zhang2022bytetrack,cao2023observation} that 
rely on the standard Kalman Filter. Our motion model has proven to be 
effective for modeling the temporal motion patterns of objects and more
robust solution than SORT-like algorithmsin non-linear and complex 
scenarios.

\textbf{SportsMOT.} To further evaluate the performance of ETTrack in non-linear scenarios, we conduct experiments on the SportsMOT benchmark. All methods used the same YOLOX detector trained on the SportsMOT training sets with or without validation sets for a fair comparison. As shown in Table \ref{sport}, these methods with $\ast$ show that their YOLOX detectors are trained on SportsMOT train and validation sets. The evaluation results presented in Table \ref{sport} show that ETTrack achieves 74.3$\%$ in HOTA, 74.5$\%$ in IDF1, 62.1$\%$ in AssA, 96.8$\%$ in MOTA, and 88.8$\%$ in DetA. Compared with MixSort\_OC\citep{cui2023sportsmot} that designs appearance based association to enhance OC\_SORT, our method still outperforms it. Specifically, ETTrack outperforms ByteTrack by up to 10.2$\%$ in HOTA and 3.1$\%$ in IDF1. Our method has shown a significant advantage in terms of object instance association, as evidenced by its performance in the HOTA and IDF1 metrics. These results indicate the effectiveness of ETTrack in dealing with non-linear motions.

\textbf{MOT17.} Table \ref{mot17} presents the tracking performance on the test set of MOT17 to validate the generalizability of the proposed motion model, which covers linear object motion. The results show that although our method achieves results comparable to existing benchmarks, it slightly underperforms relative to the current state-of-the-art methods. Given that the video sequences in MOT17 are captured under varying resolutions and diverse lighting conditions, detection emerges as a crucial element that significantly influences tracking performance. It is worth noting that although MOT17 is specifically designed to track pedestrians in scenarios where motion patterns are generally linear, our approach still achieves comparable performance with advanced methods. Thus, ETTrack consistently demonstrates robust generalizability.

\subsection{Ablation Study}

A series of ablation studies are conducted on the validation set of the DanceTrack to assess the impact of model components, momentum correction loss, and some hyper-parameters on our proposed method.

\textbf{Impact of motion modeling.} 
To assess the effectiveness of our motion predictor, we conduct a comparative study using various existing motion models. We employ diverse motion models to incorporate temporal dynamics in the tracking process, as shown in Table \ref{motion model}. Obviously, the Kalman Filter and LSTM surpass the IoU association method, demonstrating the considerable potential of motion models in tracking objects, especially when appearance information is unreliable. As shown in Table \ref{motion model}, our method has a significant advantage over the Kalman Filter, LSTM\citep{chaabane2021deft} and Vanilla Transformer\citep{Vaswani_Shazeer_Parmar_Uszkoreit_Jones_Gomez_Kaiser_Polosukhin_2017}, as measured by the HOTA and IDF1. Our motion predictor outperforms the Kalman Filter, which relies on linear motion, by up to 6.5$\%$ in HOTA and 1.9$\%$ in IDF1. Furthermore, as a deep learning-based motion predictor, our method outperforms Transformer, especially in HOTA and IDF1. Since our motion predictor Integrates TCN and Temporal Transformer models, it can learn motion patterns more comprehensively and predict object positions more accurately.

%经典三线表
\begin{table}[t]
\footnotesize
\renewcommand\arraystretch{1.3}
\setlength{\tabcolsep}{1.5pt}
\caption{Comparison of different motion models on the DanceTrack validation sets}
\label{motion model}
%标题
\centering%把表居中
\begin{tabular}{lccccc}%四个c代表该表一共四列，内容全部居中
%\begin{tabular}{@{}lccccc@{}}
\toprule%第一道横线
Tracker&HOTA$\uparrow$  &MOTA$\uparrow$ &DetA $\uparrow$  &AssA$\uparrow$  &IDF1$\uparrow$\\
\midrule%第二道横线 
No motion & 44.7   &87.3  &79.6  &25.3 &36.8\\
Kalman Filiter  &46.8    &87.5  &70.2  &31.3  &52.1 \\
LSTM  &51.2    &87.1 &76.7 &34.3    & 51.6\\
Vanilla Transformer & 51.9 & 89.3 &78.3 & 35.5 &52.7 \\
Ours & \textbf{53.3}   & \textbf{90.0}  &\textbf{78.5}  &\textbf{36.3}  &\textbf{54.0}\\
\bottomrule%第四道横线
\end{tabular}
\footnotetext{The best results are marked in bold}
\end{table}

%经典三线表
\begin{table}[t]
\footnotesize
\renewcommand\arraystretch{1.3}
\setlength{\tabcolsep}{5.2pt}
\caption{Evaluation of different model components}
\label{model component}
\centering%把表居中
\begin{tabular}{@{}lccccc@{}}
%%\begin{tabular}{l|ccccc}%四个c代表该表一共四列，内容全部居中
\toprule%第一道横线
&HOTA$\uparrow$  &MOTA$\uparrow$ &DetA $\uparrow$  &AssA$\uparrow$  &IDF1$\uparrow$\\
\midrule%第二道横线 
W/o TCN & 52.2  & 89.8 & \textbf{78.7}  &35.7 & 53.0 \\
Ours & \textbf{53.3}   & \textbf{90.0}  &78.5  &\textbf{36.3}  &\textbf{54.0}\\
\bottomrule%第四道横线
\end{tabular}
\footnotetext{"W/o" means that the TCN is removed from the motion predictor}
\footnotetext{The best results are marked in bold}
\end{table}

%经典三线表
\begin{table}[t]
\footnotesize
\renewcommand\arraystretch{1.3}
\setlength{\tabcolsep}{5.3pt}
\caption{Comparison with/without MCL }
\label{MCL}
%标题
\centering%把表居中
\begin{tabular}{@{}lccccc@{}}
%%\begin{tabular}{l|ccccc}%四个c代表该表一共四列，内容全部居中
\toprule%第一道横线
&HOTA$\uparrow$  &MOTA$\uparrow$ &DetA $\uparrow$  &AssA$\uparrow$  &IDF1$\uparrow$\\
\midrule%第二道横线 
W/o MCL & 52.5  & 89.9 & \textbf{78.6}  &35.1 & 53.1 \\
Ours & \textbf{53.3}   & \textbf{90.0}  &78.5  &\textbf{36.3}  &\textbf{54.0}\\
\bottomrule%第四道横线
\end{tabular}
\footnotetext{"W/o" means that the no motion direction information is input to motion predictor}
\footnotetext{The best results are marked in bold}
\end{table}

\textbf{Impact of model components} We conduct an ablative experiment to assess the impact of the core components on our proposed model. Specifically, the TCN is deactivated in our motion predictor to examine the resulting changes in the tracking performance. As shown in Table \ref{model component}, when the TCN is removed from the motion predictor, the HOTA and IDF1 decreased by 1.1$\%$ and 1.0$\%$, respectively. As previously discussed, TCN demonstrates strong capability for modeling short-term dependencies and commendable proficiency in capturing long-term dependencies. This capability effectively mitigates the limitations of the Temporal Transformer in modeling short-term dependencies, which are crucial for recognizing the localized features of temporal significance. The results of these experiments demonstrate that the integration of the TCN can effectively improve the prediction performance of the Temporal Transformer by extracting temporal features.

\begin{figure*}
\centering
\includegraphics[width=1\textwidth]{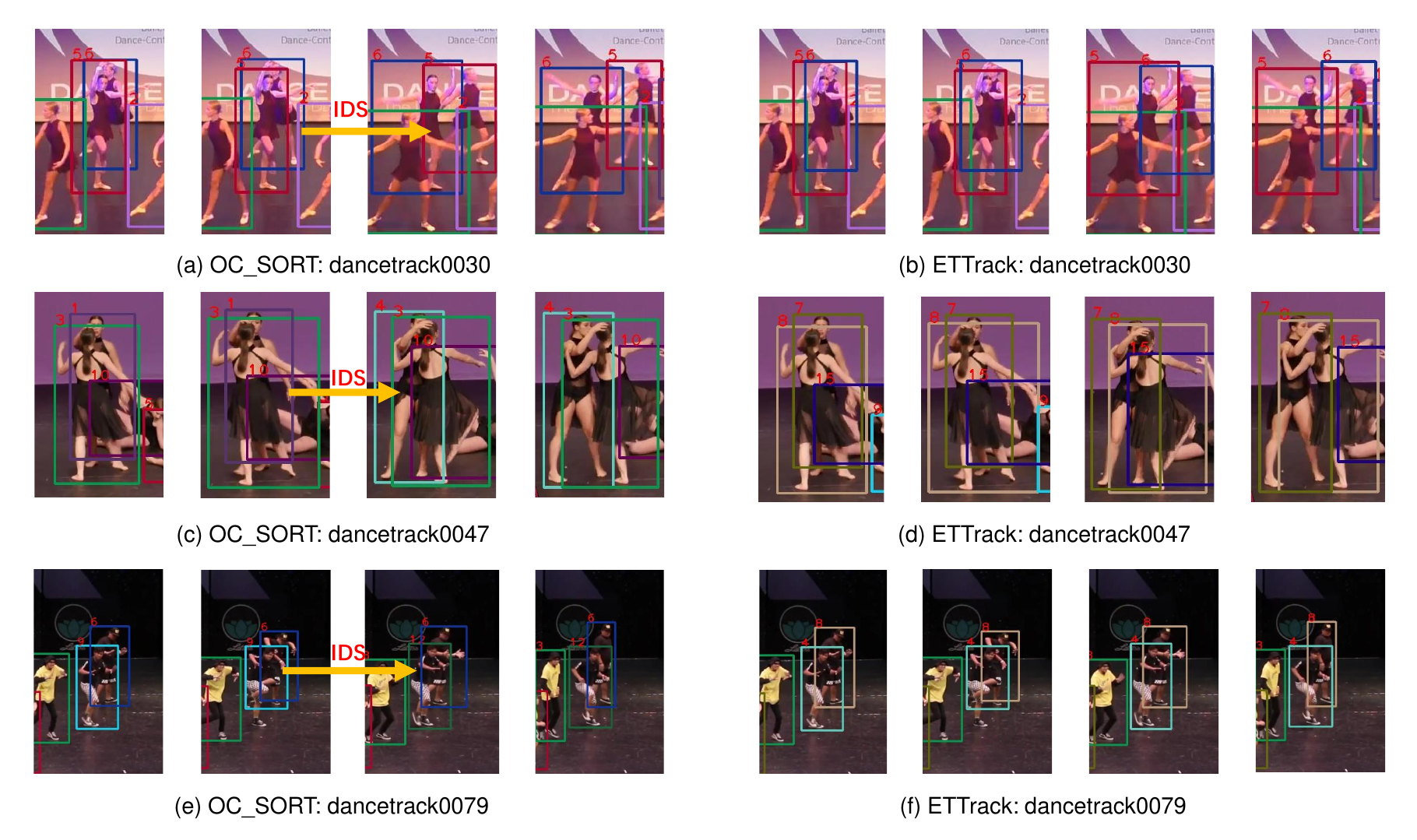}
\caption{Qualitative comparison of OC\_SORT and ETTrack (Ours). OC\_SORT leads to ID switch due to non-linear motion or severe occlusion, but ETTrack still maintains the identity. Each row shows the results' comparison for one sequence. Specifically, the problem occurs with OC\_SORT's objects: (a) ID switch between frame \#240 and \#265; (c) ID switch between frame \#315 and \#332; (f) ID switch between frame \#132 and \#146}
\label{Qualitative}
\end{figure*}

\textbf{Impact of the Momentum Correction Loss} An ablation study examines how the momentum correction loss Eq. \ref{8} behaves, as summarized by the results shown in Table \ref{MCL}. We measure the tracking performance when momentum correction loss is used to train the motion predictor. HOTA and IDF1 levels increase by 0.8$\%$ and 0.9$\%$, respectively. The results demonstrate the impact of utilizing the motion direction information in future motion prediction models. There are many sudden changes in pose and swift movements in datasets such as DanceTrack, which makes predictions that rely on past trajectories insufficient. Our future research will explore additional possibilities for incorporating more information into motion prediction.

%经典三线表
\begin{table}[t]
\footnotesize
\renewcommand\arraystretch{1.3}
\setlength{\tabcolsep}{8.2pt}
\caption{Evaluation of $p$ on the DanceTrack validation sets}
\label{p}
%标题
\centering%把表居中
\begin{tabular}{@{}lccccc@{}}
%%\begin{tabular}{l|ccccc}%四个c代表该表一共四列，内容全部居中
\toprule%第一道横线
$p$&HOTA$\uparrow$  &MOTA$\uparrow$ &DetA $\uparrow$  &AssA$\uparrow$  &IDF1$\uparrow$\\
\midrule%第二道横线 
5 & 51.5  & 89.9 & 77.8  &35.1 & 52.1 \\
8  &53.1 &89.9 &78.2 &35.5  &53.4   \\
10 & \textbf{53.3}   & \textbf{90.0}  &\textbf{78.5}  \textbf&{36.3}  &\textbf{54.0}\\
13  &53.2  &90.0  &78.4  &\textbf{36.4}  &53.8 \\
15  &52.9  &90.1  &78.1  &36.2  &53.1 \\
\bottomrule%第四道横线
\end{tabular}
\footnotetext{The best results are marked in bold}
\end{table}

%经典三线表
\begin{table}[t]
\footnotesize
\renewcommand\arraystretch{1.3}
\setlength{\tabcolsep}{8pt}
%%\footnotesize
\caption{Evaluation of $\beta$ in Eq. \ref{8}}
%标题
\label{beta}
\centering%把表居中
\begin{tabular}{@{}lccccc@{}}
%%\begin{tabular}{l|ccccc}%四个c代表该表一共四列，内容全部居中
\toprule%第一道横线
$\beta$&HOTA$\uparrow$  &MOTA$\uparrow$ &DetA $\uparrow$  &AssA$\uparrow$  &IDF1$\uparrow$\\
\midrule%第二道横线 
0 & 52.2  & 89.9 & 78.7  &35.5 & 53.1 \\
0.1  &52.6 &89.9 &78.6 &35.6  &53.3   \\
0.2  &52.9  &\textbf{90.1}  &\textbf{78.9}  &36.0  &53.7 \\
0.3 & \textbf{53.3}   & 90.0  &78.5  &\textbf{36.3}  &\textbf{54.0}\\
0.4  &52.7  &89.8  &78.5  &36.2  &53.5 \\
0.5  &51.7  &89.8 &78.6  &35.6  &52.9  \\
\bottomrule%第四道横线
\end{tabular}
\footnotetext{The best results are marked in bold}
\end{table}

\begin{figure*}
\centering
\includegraphics[width=1\textwidth]{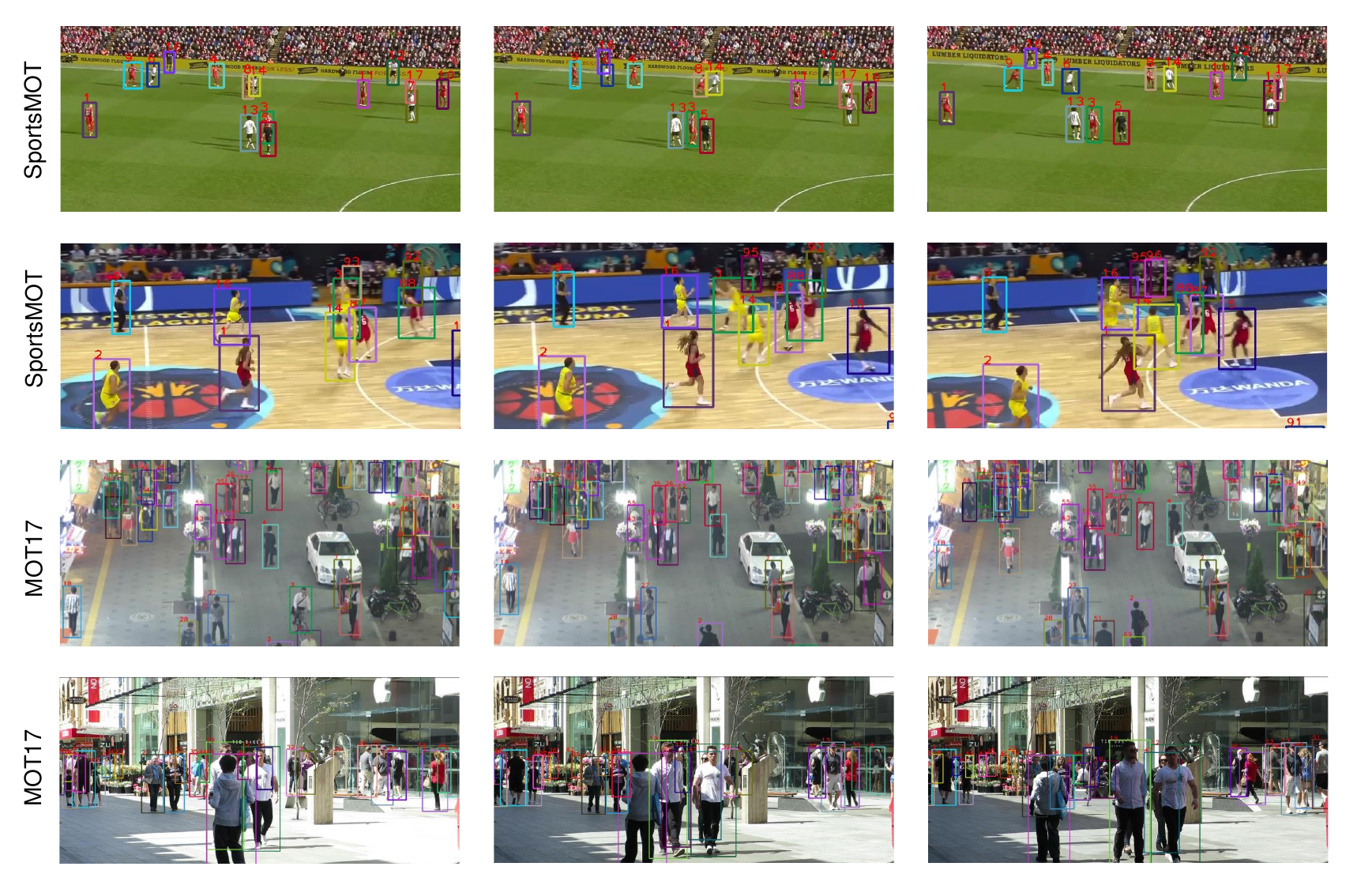}
\caption{ The visualizations of ETTrack's tracking results on the test set of SportsMOT and MOT17. Boxes of the same color denote the same object
}
\label{Qualitative1}
\end{figure*}

\textbf{Impact of historical trajectory length} To demonstrate how the tracking performance is affected by the length of the historical trajectory, we evaluate our proposed method at different $p$ values. The results, as listed in Table \ref{p}, indicate that a very small historical trajectory length fails to provide sufficient information, resulting in unreliable predictions. Our results indicate that extending the historical trajectory length provides a more comprehensive analysis of object motion. However, a very large historical trajectory length tends to provide a considerable degree of noise, which in turn negatively affects the tracking performance. Therefore, we selected $T$ to be 10, which accounts for 0.5 seconds of the object's historical trajectory, based on a video frame rate of 20 FPS. The results presented in Table \ref{p} demonstrate the significance of choosing an appropriate historical trajectory length to achieve superior performance in object-tracking tasks.

\textbf{Impact of the weight of the Momentum Correction Loss} Finally, we also explore the effect of the hyper-parameter $\beta$, which determines the degree to which the momentum correction loss affects the final objective function. As shown in Table \ref{beta}, the best results are obtained on the DanceTrack validation sets when $\beta$ was set to 0.3.

\subsection{Qualitative Results}

Fig. \ref{Qualitative} shows qualitative comparisons of ETTrack and OC\_SORT. The first row shows that OC\_SORT causes ID switching owing to the object occlusion or non-linear motion. The Kalman Filter's assumption of linear motion prevents OC\_SORT from accurately predicting sudden pose changes, resulting in false matches. By contrast, ETTrack maintains consistent identities and exhibits robustness in handling non-linear motions. This demonstrates that our method can accurately predict the objects’ position when the objects exhibit complex and non-linear motions. 

The visualized results of ETTrack on the test set of SportsMOT and MOT17 are shown in Fig. \ref{Qualitative1}. ETTrack provides accurate predictions on the test set of SportsMOT. It is demonstrated that our method can predict the positions of objects accurately in sports scenarios where the objects exhibit rapid and non-linear motions. Fig. \ref{Qualitative1} also shows several ETTrack's tracking results on the test set of MOT17. It can be observed that although the MOT17 dataset is designed to track pedestrians in scenarios where motion patterns are generally linear, our method still delivers impressive tracking results.

\section{Conclusion}

In this paper, we propose a motion-based MOT called ETTrack, which uses an enhanced temporal motion predictor to improve object association and tracking performance in non-linear motion. The motion predictor integrates a Temporal Transformer model and a Temporal Convolutional Network (TCN) to capture both local and global historical motion information. In addition, the proposed method uses a novel Momentum Correction Loss to guide the motion predictor during training and improve its ability to handle complex movements. As a result, compared to other motion models based on the Kalman Filter and deep learning, ETTrack exhibits better prediction performance on challenging datasets such as DanceTrack and SportsMOT. Simultaneously, it achieves comparable performance on pedestrian-centric datasets such as MOT17. In future work, we will conduct further research on camera motion information and human pose features in the motion model.

\backmatter

\bmhead*{Funding}
The authors did not receive support from any organization for the submitted work.

\section*{Declarations}

\bmhead*{Conflict of interest}
Not applicable.
\bmhead*{Data availability}
All datasets used are publicly available.

%%===========================================================================================%%
%% If you are submitting to one of the Nature Portfolio journals, using the eJP submission   %%
%% system, please include the references within the manuscript file itself. You may do this  %%
%% by copying the reference list from your .bbl file, paste it into the main manuscript .tex %%
%% file, and delete the associated \verb+\bibliography+ commands.                            %%
%%===========================================================================================%%

\bibliography{sn-bibliography}% common bib file

%% if required, the content of .bbl file can be included here once bbl is generated
%%\input sn-article.bbl

\end{document}